\documentclass[final]{cvpr}

\usepackage{times}
\usepackage{epsfig}
\usepackage{graphicx}
\usepackage{amsmath}
\usepackage{amssymb}

\usepackage[pagebackref=true,breaklinks=true,colorlinks,bookmarks=false]{hyperref}

\def\cvprPaperID{8} 

\def\cvprPaperID{8} 
\def\confYear{CVPR 2021}

\begin{document}

\setlength{\abovedisplayskip}{2.5pt}
\setlength{\belowdisplayskip}{2.5pt}

\title{\vspace{-6em}PC-DAN: Point Cloud based Deep Affinity Network for 3D Multi-Object Tracking\vspace{-1.4em}}  
\author{Aakash Kumar*, Jyoti Kini*, Mubarak Shah*, Ajmal Mian†\\
*Center for Research in Computer Vision, University of Central Florida, Orlando, FL, USA\\
†University of Western Australia, Crawley, WA, Australia\vspace{-1.8em}\\
}
\maketitle

In recent times, the scope of LIDAR (Light Detection and Ranging) sensor-based technology has spread across numerous fields.  It is popularly used to map terrain and navigation information into reliable 3D point cloud data, potentially revolutionizing the autonomous vehicles and assistive robotic industry. A point cloud is a dense compilation of spatial data in 3D coordinates. It plays a vital role in modeling complex real-world scenes since it preserves structural information and avoids perspective distortion, unlike image data, which is the projection of a 3D structure on a 2D plane. In order to leverage the intrinsic capabilities of the LIDAR data, we propose a PointNet-based approach for 3D Multi-Object Tracking (MOT).

\begin{figure}[h]
\begin{center}
  \includegraphics[scale=0.25]{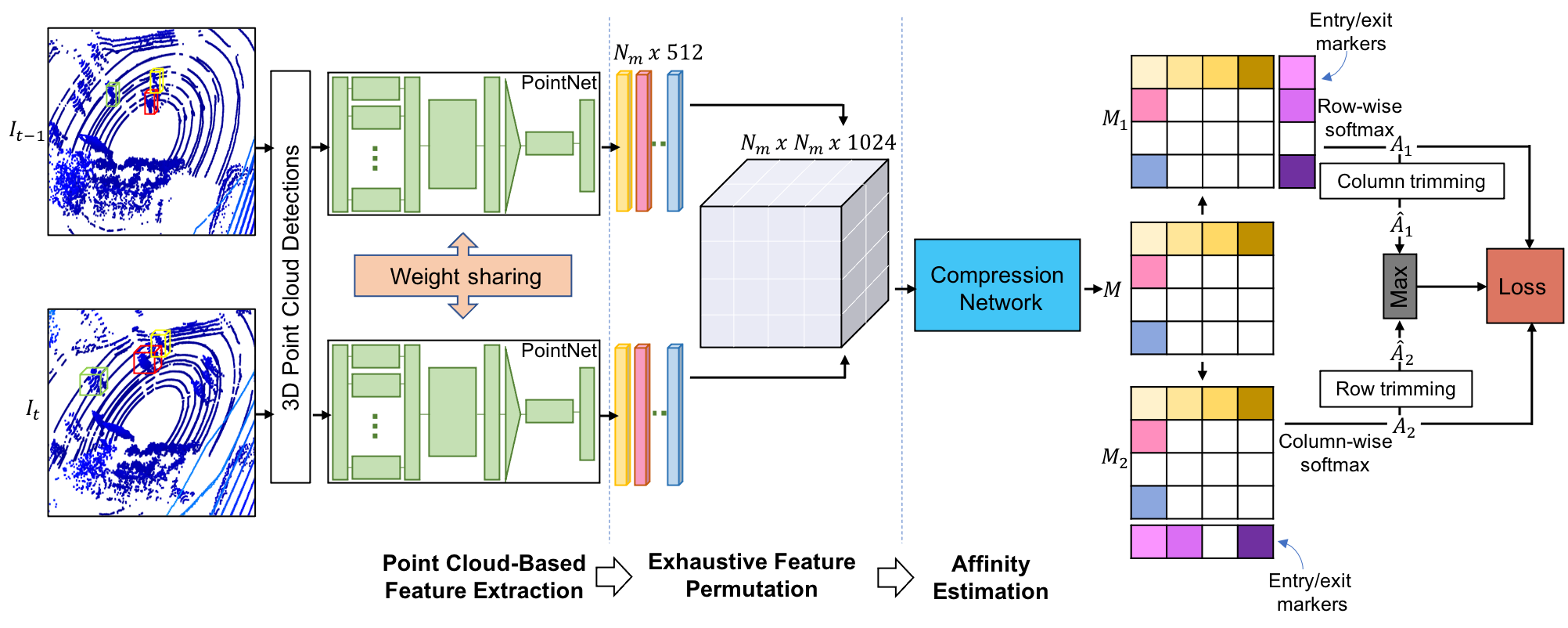}
\end{center}
   \caption{PC-DAN Architecture}
\label{fig:Architecture}
\end{figure}

Given the 3D detections, our proposed model predicts the affinity between two frames. Our network architecture derives inspiration from Deep Affinity Network (DAN) \cite{sun2019deep}, which focuses on 2D tracking using image data. We replace VGG-16 in DAN architecture with PointNet \cite{qi2017pointnet} for feature extraction from LIDAR data and utilize the affinity estimator for 3D MOT as shown in Figure \ref{fig:Architecture}. Our solution consists of a PointNet feature extractor, a component for exhaustive pairing permutation of features, and a compression network. We, initially, crop objects from the point cloud using 3D detections and pass these through PointNet to generate the features for each object.  Exhaustive permutations of these feature vectors $F_t$ and $F_{t-n}$ are encoded in a tensor $ \in R^{N_{m} \times N_{m} \times 1024}$, where $N_m$ is the number of objects in each frame. For our experiments, we set $N_m$ to 100 objects. Next, a compression network  consisting of 5 convolution layers is used to map the encodings to $M \in R^{N_{m} \times N_{m}}$. The resultant matrix $M$ is augmented to $M_1$ and $M_2$ matrices by appending an extra column and row respectively to account for the objects entering and leaving the scene.  Thereafter, row-wise and column-wise softmax operations are performed to generate corresponding $A_1$ and $A_2$ matrices. These matrices, along with their column and row trimmed versions $\hat{A}_1$ and $\hat{A}_2$ respectively, are used for the loss computation. $A_1$ and $A_2$ matrices are used to predict the affinity between a pair of frames. After the prediction score is obtained from the network, we use a linear programming solver to find the optimal tracks from the predicted affinity scores.
 
In addition, we use the forward-direction loss, backward-direction loss, consistency loss, and assemble loss functions from DAN to train our model. Here, $G$ is the ground-truth affinity. $G_1$ are $G_2$ are trimmed versions of ground-truth constructed after ignoring the last row and column respectively, which encompasses objects entering and leaving the scene. $G_3$ is trimmed version generated by ignoring both the last row and column. Forward-direction loss $L_f$ is used to learn the identity association from previous frame $I_{t-n}$ to the current frame $I_t$. 
\begin{align}
    L_{f} (G_1,A_1) = \frac{\sum (G_1 \odot (-\log{A_1}))}{\sum G_1}
\end{align}
Likewise, backward-direction loss $L_b$ encourages identity association from current frame $I_t$ to previous frame $I_{t-n}$. 
\begin{align}
    L_{b} (G_2,A_2) = \frac{\sum (G_2 \odot (-\log{A_2}))}{\sum G_2}
\end{align}
The consistency loss $L_c$ mitigates the disagreement between forward and backward loss.
\begin{align}
    L_{c} (\hat{A}_1,\hat{A}_2) = ||\hat{A}_1 - \hat{A}_2||_1
\end{align}
The assemble loss $L_a$ suppresses non-maximal forward and backward associations.
\begin{align}
    L_{a} (G_3,\hat{A}_1,\hat{A}_2) = \frac{\sum (G_3 \odot (-\log{(\max(\hat{A}_2,\hat{A}_1)})))}{\sum G_3}
\end{align}
The overall loss $L_a$ constitutes the above four losses and is given by:
\begin{align}
    L = \frac{L_f + L_b + L_c + L_a}{4}
\end{align}

For experimental evaluations, we use the pre-trained weights from the Kitti dataset and fine-tune the model on Jackrabot Dataset (JRDB). Using the validation split recommended on the JRDB website, we have achieved 88.06\% MOTA on validation data. From test data, we use all the detections with a confidence score greater than 0.4 to account for a large number of spurious detections. Currently, our submission, titled {\bf P-DAN\_CVPR2021}, stands at the top of the leaderboard on test data with MOTA 22.56\%. Table \ref{table:Results} summarises the quantitative analysis on the JRDB dataset.

\setlength{\tabcolsep}{2.2pt}  
\begin{table}[h]
\renewcommand{\arraystretch}{1.06}
\begin{center}
\begin{tabular}{|c|c|c|c|c|c|c|}
\hline
Name    &	MOTA    &	MOTP    &	IDs &	False   &	False   &	Time \\
        &	    &	    &	 &	+ve   &	-ve   &	(s) \\
\hline\hline
Ours    &	{\bf 22.56}   &	6.03    &	26009   &	58090   &	681852  &	0.16 \\
\hline
JRMOT \cite{shenoi2020jrmot}   &	20.15   &	42.46   &	4207    &	19711   &	765907  &	0.06 \\
\hline
AB3DMOT \cite{weng20203d} &	19.35   &	42.02   &	6177    &	13664  &	777946  &	0.01 \\
\hline
\end{tabular}
\end{center}
\caption{Quantitative analysis of our method}
\label{table:Results}
\end{table}

{\small
\bibliographystyle{ieee_fullname}
\bibliography{egbib}
}

\end{document}